# Continuous Representation of Location for Geolocation and Lexical Dialectology using Mixture Density Networks

**Afshin Rahimi**     **Timothy Baldwin**     **Trevor Cohn**
School of Computing and Information Systems
The University of Melbourne
arahimi@student.unimelb.edu.au
{tbaldwin,t.cohn}@unimelb.edu.au

## Abstract

We propose a method for embedding two-dimensional locations in a continuous vector space using a neural network-based model incorporating mixtures of Gaussian distributions, presenting two model variants for text-based geolocation and lexical dialectology. Evaluated over Twitter data, the proposed model outperforms conventional regression-based geolocation and provides a better estimate of uncertainty. We also show the effectiveness of the representation for predicting words from location in lexical dialectology, and evaluate it using the DARE dataset.

## 1 Introduction

Geolocation is an essential component of applications such as traffic monitoring (Emadi et al., 2017), human mobility pattern analysis (McNeill et al., 2016; Dredze et al., 2016) and disaster response (Ashktorab et al., 2014; Wakamiya et al., 2016), as well as targeted advertising (Anagnostopoulos et al., 2016) and local recommender systems (Ho et al., 2012). Although Twitter provides users with the means to geotag their messages, less than 1% of users opt to turn on geotagging, so third-party service providers tend to use profile data, text content and network information to infer the location of users. Text content is the most widely used source of geolocation information, due to its prevalence across social media services.

Text-based geolocation systems use the geographical bias of language to infer the location of a user or message using models trained on geotagged posts. The models often use a representation of text (e.g. based on a bag-of-words, convolutional or recurrent model) to predict the location either in real-valued latitude/longitude coordinate space or in discretised region-based space, using regression or classification, respectively. Regression models, as a consequence of minimising squared loss for a unimodal distribution, predict inputs with multiple targets to lie between the targets (e.g. a user who mentions content in both NYC and LA is predicted to be in the centre of the U.S.). Classification models, while eliminating this problem by predicting a more granular target, don't provide fine-grained predictions (e.g. specific locations in NYC), and also require heuristic discretisation of locations into regions (e.g. using clustering).

Mixture Density Networks ("MDNs": Bishop (1994)) alleviate these problems by representing location as a mixture of Gaussian distributions. Given a text input, an MDN can generate a mixture model in the form of a probability distribution over all location points. In the example of a user who talks about both NYC and LA, e.g., the model will predict a strong Gaussian component in NYC and another one in LA, and also provide an estimate of uncertainty over all the coordinate space.

Although MDNs are not new, they have not found widespread use in *inverse* regression problems where for a single input, multiple correct outputs are possible. Given the integration of NLP technologies into devices (e.g. phones or robots with natural language interfaces) is growing quickly, there is a potential need for interfacing language with continuous variables as input or target. MDNs can also be used in general text regression problems such as risk assessment (Wang and Hua, 2014), sentiment analysis (Joshi et al., 2010) and loan rate prediction (Bitvai and Cohn, 2015), not only to improve prediction but also to use the mixture model as a representation for the continuous variables. We apply MDNs to geotagged Twitter data in two different settings: (a) predicting location given text; and (b) predicting text given location.

Geotagged text content is not only useful in geolocation, but can also be used in lexical dialectology. Lexical dialectology is (in part) the converse of text-based geolocation (Eisenstein, 2015): instead of predicting location from language, language (e.g. dialect terms) are predicted from a given location. This is a much more challenging task as the lexical items are not known beforehand, and there is no notion of dialect regions in the continuous space of latitude/longitude coordinates. A lexical dialectology model should not only be able to predict dialect terms but also be able to automatically learn dialect regions.

In this work, we use bivariate Gaussian mixtures over geotagged Twitter data in two different settings, and demonstrate their use for geolocation and lexical dialectology. Our contributions are as follows: (1) we propose a continuous representation of location using bivariate Gaussian mixtures; (2) we show that our geolocation model outperforms regression-based models and achieves comparable results with classification models, but with added uncertainty over the continuous output space; (3) we show that our lexical dialectology model is able to predict geographical dialect terms from latitude/longitude input with state-of-the-art accuracy; and (4) we show that the automatically learned Gaussian regions match expert-generated dialect regions of the U.S.[1]

## 2 Related Work

### 2.1 Text-based Geolocation

Text-based geolocation models are defined as either a regression or a classification problem. In regression geolocation, the model learns to predict a real-valued latitude/longitude from a text input. This is a very challenging task for data types such as Twitter, as they are often heavily biased toward population centres and urban areas, and far from uniform. As an example, *Norwalk* is the name of a few cities in the U.S among which Norwalk, California (West Coast) and Norwalk, Connecticut (East Coast) are the two most populous cities. Assuming that occurrences of the city's name are almost equal in both city regions within the training set, a trained regression-based geolocation model given *Norwalk* as input, would geolocate it to a point in the middle of the U.S. instead of choosing one of the cities. In the machine learning literature, regression problems where there are multiple real-valued outputs for a given input are called *inverse problems* (Bishop, 1994). Here, standard regression models predict an average point in the middle of all training target points to minimise squared error loss. Bishop (1994) proposes density mixture networks to model such inverse problems, as we discuss in detail in Section 3.

In addition, non-Bayesian interpretations of regression models, which are often used in practice, don't produce any prediction of uncertainty, so other than the predicted point, we have little idea where else the term could have high or low probability. Priedhorsky et al. (2014) propose a Gaussian Mixture Model (GMM) approach instead of squared loss regression, whereby they learn a mixture of bivariate Gaussian distributions for each individual $n$-gram in the training set. During prediction, they add the Gaussian mixture of each $n$-gram in the input text, resulting in a new Gaussian mixture which can be used to predict a coordinate with associated uncertainty. To add the mixture components they use a weighted sum, where the weight of each $n$-gram is assigned by several heuristic features. Learning a GMM for each $n$-gram is resource-intensive if the size of the training set — and thus the number of $n$-grams — is large.

Assuming sufficient training samples containing the term *Norwalk* in the two main, a trained classification model would, given this term as input, predict a probability distribution over all regions, and assign higher probabilities to the regions containing the two major cities. The challenge, though, is that the coordinates in the training data must first be partitioned into regions using administrative regions (Cheng et al., 2010; Hecht et al., 2011; Kinsella et al., 2011; Han et al., 2012, 2014), a uniform grid (Serdyukov et al., 2009), or a clustering method such as a $k$-d tree (Wing and Baldridge, 2011) or $K$-means (Rahimi et al., to appear). The cluster/region labels can then be used as targets. Once we have a prediction about where a user is more likely to be from, there is no more information about the coordinates inside the predicted region. If a region that contains Wyoming is predicted as the home location of a user, we have no idea which city or county within Wyoming the user might be from, unless we retrain the model using a more fine-grained discretisation or a hierarchical discretisation (Wing and Baldridge, 2014), which is both time-consuming and challenging due to data

---
[1] Code available at
https://github.com/afshinrahimi/geomdn

sparseness.

## 2.2 Lexical Dialectology

The traditional linguistic approach to lexical dialectology is to find the geographical distributions of known contrast sets such as {*you, yall, yinz*}: (Labov et al., 2005; Nerbonne et al., 2008; Gonçalves and Sánchez, 2014; Doyle, 2014; Huang et al., 2015; Nguyen and Eisenstein, to appear). This usually involves surveying a large geographically-uniform sample of people from different locations and analysing where each known alternative is used more frequently. Then, the coordinates are clustered heuristically into dialect regions, based on the lexical choices of users in each region relative to the contrast set. This processing is very costly and time-consuming, and relies critically on knowing the lexical alternatives a priori. For example, it would require a priori knowledge of the fact that people in different regions of the US use *pop* and *soda* to refer to the same type of drink, and a posteriori analysis of the empirical geographical distribution of the different terms. Language, particularly in social media and among younger speakers, is evolving so quickly, in ways that can be measured over large-scale data samples such as Twitter, that we ideally want to be able to infer such contrast sets dynamically. The first step in automatically collecting dialect words is to find terms that are disproportionately distributed in different locations. The two predominant approaches to this problem are model-based (Eisenstein et al., 2010; Ahmed et al., 2013; Eisenstein, 2015) and through the use of statistical metrics (Monroe et al., 2008; Cook et al., 2014). Model-based approaches use a topic model, e.g., to extract region-specific topics, and from this, predict the probability of seeing a word given a geographical region (Eisenstein et al., 2010). However, there are scalability issues, limiting the utility of such models.

In this paper, we propose a neural network architecture that learns a mixture of Gaussian distributions as its activation function, and predicts both locations from word-based inputs (geolocation), and words from location-based inputs (lexical dialectology).

## 3 Model

### 3.1 Bivariate Gaussian Distribution

A bivariate Gaussian distribution is a probability distribution over 2d space (in our case, a latitude/longitude coordinate pair). The probability mass function is given by:

$$\mathcal{N}(\boldsymbol{x}|\boldsymbol{\mu}, \Sigma) = \frac{1}{(2\pi)} \frac{1}{|\Sigma|^{1/2}}$$
$$\exp\left\{-\frac{1}{2}(\boldsymbol{x}-\boldsymbol{\mu})^{\intercal}\Sigma^{-1}(\boldsymbol{x}-\boldsymbol{\mu})\right\}$$

where $\boldsymbol{\mu}$ is the 2-dimensional mean vector, the matrix $\Sigma = \begin{pmatrix} \sigma_1^2 & \rho_{12}\sigma_1\sigma_2 \\ \rho_{12}\sigma_1\sigma_2 & \sigma_2^2 \end{pmatrix}$ is the covariance matrix, and $|\Sigma|$ is its determinant. $\sigma_1$ and $\sigma_2$ are the standard deviations of the two dimensions, and $\rho_{12}$ is the covariance. $\boldsymbol{x}$ is a latitude/longitude coordinate whose probability we are seeking to predict.

### 3.2 Mixtures of Gaussians

A mixture of Gaussians is a probabilistic model to represent subpopulations within a global population in the form of a weighted sum of $K$ Gaussian distributions, where a higher weight with a component Gaussian indicates stronger association with that component. The probability mass function of a Gaussian mixture model is given by:

$$\mathcal{P}(\boldsymbol{x}) = \sum_{k=1}^{K} \pi_k \mathcal{N}(\boldsymbol{x}|\boldsymbol{\mu_k}, \Sigma_k)$$

where $\sum_{k=1}^{K} \pi_k = 1$, and the number of components $K$ is a hyper-parameter.

### 3.3 Mixture Density Network (`MDN`)

A mixture density network ("MDN": Bishop (1994)) is a latent variable model where the conditional probability of $p(y|x)$ is modelled as a mixture of $K$ Gaussians where the mixing coefficients $\pi$ and the parameters of Gaussian distributions $\boldsymbol{\mu}$ and $\Sigma$ are computed as a function of input using a neural network:

$$\mathcal{P}(y|x) = \sum_{k=1}^{K} \pi_k(x) \mathcal{N}\big(y|\mu_k(x), \Sigma_k(x)\big)$$

In the bivariate case of latitude/longitude, the number of parameters of each Gaussian is 6 ($\pi_k(x), \mu_{1k}(x), \mu_{2k}(x), \rho_k(x), \sigma_{1k}(x), \sigma_{2k}(x)$), which are learnt in the output layer of a regular neural network as a function of input $x$. The output size of the network for $K$ components would be $6 \times K$. The output of an MDN for $N$ samples (e.g. where $N$ is the mini-batch size) is an $N \times 6K$ matrix which is then sliced and reshaped into ($N \times 2 \times K$), ($N \times 2 \times K$), ($N \times 1 \times K$) and ($N \times 1 \times K$) matrices, providing the model parameters $\mu$, $\sigma$, $\rho$ and $\pi$. Each parameter type has its own constraints: $\sigma$ (the standard deviation) should be positive, $\rho$ (the correlation) should be in the interval $[-1, 1]$ and $\pi$ should be positive and sum to one, as a probability distribution. To force these constraints, the following transformations are often applied to each parameter set:

$$\sigma \sim \text{SoftPlus}(\sigma\prime) = \log(\exp(\sigma\prime) + 1) \in (0, +\infty)$$
$$\pi \sim \text{SoftMax}(\pi\prime)$$
$$\rho \sim \text{SoftSign}(\rho\prime) = \frac{\rho\prime}{1 + |\rho\prime|} \in [-1, 1]$$

As an alternative, it's possible to use transformations like $\exp$ for $\sigma$ and $\tanh$ for $\rho$. After applying the transformations to enforce the range constraints, the negative log likelihood loss of each sample $x$ given a 2d coordinate label $y$ is computed as:

$$\mathcal{L}(y|x) = -\log\left\{\sum_{k=1}^{K} \pi_k(x)\mathcal{N}(y|\mu_k(x), \Sigma_k(x))\right\}$$

To predict a location, given an unseen input, the output of the network is reshaped into a mixture of Gaussians and $\mu_k$, one of the $K$ components' $\mu$ is chosen as the prediction. The selection criteria is either based on the strongest component with highest $\pi$, or the component that maximises the overall mixture probability:

$$\max_{\mu_i \in \{\mu_1...\mu_K\}} \sum_{k=1}^{K} \pi_k \mathcal{N}(\mu_i|\mu_k, \Sigma_k)$$

For further details on selection criteria, see Bishop (1994).

### 3.4 Mixture Density Network with Shared Parameters (`MDN-SHARED`)

In the original MDN model proposed by Bishop (1994), the parameters of the mixture model are separate functions of input, which is appropriate when the inputs and outputs directly relate to each other, but in the case of geolocation or lexical dialectology, the relationship between inputs and outputs is not so obvious. As a result, it might be a difficult task for the model to learn all the parameters of each sample correctly. Instead of using the output to predict all the parameters, we share $\mu$ and $\Sigma$ among all samples as parameters of the output layer, and only use the input to predict $\pi$, the mixture probabilities, using a SoftMax layer. We initialise $\mu$ by applying $K$-means clustering to the training coordinates and setting each value of $\mu$ to the centroids of the $K$ clusters; we initialise $\Sigma$ randomly between 0 and 10. We use the original cost function to update the weight matrices, biases and the global shared parameters of the mixture model through backpropagation. Prediction is performed in the same way as for MDN.

### 3.5 Continuous Representation of Location

Gaussian mixtures are usually used as the output layer in neural networks (as in MDN) for inverse regression problems. We extend their application by using them as an input representation when the input is a multidimensional continuous variable. In problems such as lexical dialectology, the input is real-valued 2d coordinates, and the goal is to predict dialect words from a given location. Small differences in latitude/longitude may result in big shifts in language use (e.g. in regions such as Switzerland or Gibraltar). One way to model this is to discretise the input space (similar to the discretisation of the output space in classification), with the significant downside that the model is not able to learn/fine-tune regions in a data-driven way. A better solution is to use a $K$ component Gaussian mixture representation of location, where $\mu$ and $\Sigma$ are shared among all samples, and the output of the layer is the probability of input in each of the mixture components. Note that in this representation, there is no need for $\pi$ parameters as we just need to represent the association of an input location to $K$ regions, which will then be used as input to the next layer of a neural network and used to predict the targets. We use this continuous representation of location to predict dialect words from location input.

## 4 Experiments

We apply the two described MDN models on two widely-used geotagged Twitter datasets for geolocation, and compare the results with state-of-the-art classification and regression baselines. Also, we use the mixture of Gaussian representation of location to predict dialect terms from coordinates.

### 4.1 Data

In our experiments, we use two existing Twitter user geolocation datasets: (1) GEOTEXT (Eisenstein et al., 2010), and (2) TWITTER-US (Roller et al., 2012). Each dataset has fixed training, devel-opment and test partitions, and a user is represented by the concatenation of their tweets, and labelled with the latitude/longitude of the first collected geo-tagged tweet.[2] GEOTEXT and TWITTER-US cover the continental US with 9k, 449k users, respectively.[3]

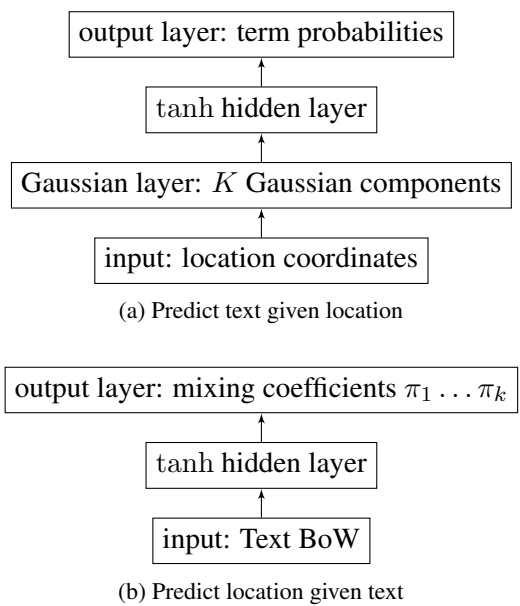

Figure 1: (a) The lexical dialectology model using a Gaussian representation layer. (b) `MDN-SHARED` geolocation model where the mixture weights $\pi$ are predicted for each sample, and $\mu$ and $\Sigma$ are parameters of the output layer, shared between all samples.

**DARE** is a dialect-term dataset derived from the Dictionary of American Regional English (Cassidy, 1985) by Rahimi et al. (to appear). DARE consists of dialect regions, terms and the meaning of each term.[4] It represents the aggregation of a number of dialectal surveys over different regions of the U.S., to identify shared dialect regions. Because the dialect regions in DARE maps are not machine readable, populous cities within each dialect region are manually extracted and associated with their dialect region terms. The dataset is made up of around 4.3k dialect terms across 99 U.S. dialect regions.

---

[2]This geolocation representation is naive, but was made by the creators of the original datasets and has been used by others. It has been preserved in this work for comparability with the results of others, despite misgivings about whether this is a faithful representation of the location for a given user.
[3]The datasets can be obtained from https://github.com/utcompling/textgrounder.
[4]http://www.daredictionary.com/

## 4.2 Geolocation

We use a 3-layer neural network as shown in Figure 1a where the input is the $l_2$ normalised bag-of-words model of a given user with stop words, @-mentions and words with document frequency less than 10 removed. The input is fed to a hidden layer with `tanh` nonlinearity that produces the output of the network (with no nonlinearity applied). The output is the collection of Gaussian mixture parameters $(\mu, \Sigma, \pi)$ from `MDN`. For prediction, the $\mu_k$ of the mixture component which has the highest probability within the mixture component is selected. In the case of `MDN-SHARED`, the output is only $\pi$, a vector with size $K$, but the output layer contains extra global parameters $\mu$ and $\Sigma$ ($\sigma_{\text{lat}}, \sigma_{\text{lon}}, \rho$) which are shared between all the samples. The negative log likelihood objective is optimised using Adam (Kingma and Ba, 2014) and early stopping is used to prevent overfitting. The hidden layer is subject to drop-out and elastic net regularisation (with equal $l_1$ and $l_2$ shares). As our baseline, we used a multilayer perceptron regressor with the same input and hidden architecture but with a 2d output with linear activation that predicts the location from text input. The regularisation and drop-out rate, hidden layer size and the number of Gaussian components $K$ (for `MDN` and `MDN-SHARED`) are tuned over the development set of each dataset, as shown in Table 1.

We evaluate the predictions of the geolocation models based on three measures (following Cheng et al. (2010) and Eisenstein et al. (2010)):

1. the classification accuracy within a 161km (= 100 mile) radius of the actual location ("Acc@161"); i.e., if the predicted location is within 161km of the actual location, it is considered to be correct
2. the mean error ("Mean") between the predicted location and the actual location of the user, in kilometres
3. the median error ("Median") between the predicted location and the actual location of the user, in kilometres

## 4.3 Lexical Dialectology

To predict dialect words from location, we use a 4-layer neural network as shown in Figure 1b. The input is a latitude/longitude coordinate, the first hidden layer is a Gaussian mixture with $K$ components which has $\mu$ and $\Sigma$ as its parameters and produces a probability for each component as an activation function, the second hidden layer with tanh nonlinearity captures the association between different Gaussians, and the output is a SoftMax layer which results in a probability distribution over the vocabulary. For a user label, we use an $l_1$ normalised bag-of-words representation of its text content and use binary $tf$ and $idf$ for term-weighting. The model should learn to predict the probability distribution over the vocabulary and so be capable of predicting dialect words with a higher probability. It also learns regions (parameters of $K$ Gaussians) that represent dialect regions.

We evaluate the lexical dialectology model (`MDN-layer`) using perplexity of the predicted unigram distribution, and compare it with a baseline where the Gaussian mixture layer is replaced with a tanh hidden layer (`tanh-layer`). Also we retrieve words given points within a region from the DAREDS dataset, and measure recall with respect to relevant dialect terms from DAREDS. To do that, we randomly sample $P = 10000$ latitude/longitude points from the training set and predict the corresponding word distribution. To come up with a ranking over words given region $r$ as query, we use the following measure:

$$score(w_i|r) = \frac{1}{N} \sum_{p_j \in r} \log(P(w_i|p_j))$$
$$- \frac{1}{P} \sum_{j=1}^{P} \log(P(w_i|p_j))$$

where $N$ equals the number of points (out of 10000) inside the query dialect region $r$ and $P$ equals the total number of points (here 10000). For example, if we are querying dialect terms from dialect region South ($r$), $N$ is the number of randomly selected points that fall within the constituent states of South. $score(w_i|r)$ measures the (log) probability ratio of a word $w_i$ inside region $r$ compared to its global score: if a word is local to region $r$, the ratio will be higher. We use this measure to create a ranking over the vocabulary from which we measure precision and recall at $k$ given gold-standard dialect terms in DAREDS.

## 5 Results

### 5.1 Geolocation

The performance of `Regression`, `MDN` and `MDN-SHARED`, along with several state-of-the-art classification models, is shown in Table 2. The `MDN` and `MDN-SHARED` models clearly outperform `Regression`, and achieve competitive or slightly worse results than the classification models but provide uncertainty over the whole output space. The geographical distribution of error for `MDN-SHARED` over the development set of TWITTER-US is shown in Figure 3, indicating larger errors in MidWest and particularly in North Pacific regions (e.g. Oregon).

### 5.2 Dialectology

The perplexity of the lexical dialectology model using Gaussian mixture representation (`MDN-layer`) is 840 for the 54k vocabulary of TWITTER-US dataset, 1% lower than a similar network architecture with a tanh hidden layer (`tanh-layer`), which is not a significant improvement. Also we evaluated the model using recall at $k$ and compared it to the `tanh-layer` model which again is competitive with `tanh-layer` but with the advantage of learning dialect regions simultaneously. Because the DARE dialect terms are not used frequently in Twitter, many of the words are not covered in our dataset, despite its size. However, our model is able to retrieve dialect terms that are distinctly associated with regions. The top dialect words for regions New York, Louisiana, Illinois and Pennsylvania are shown in Table 3, and include named entities, dialect words and local hashtags. We also visualised the learned Gaussians of the dialectology model in Figure 2, which as expected show several smaller regions (Gaussians with higher $\sigma$) and larger regions in lower populated areas. It is interesting to see that the shape of the learned Gaussians matches natural borders such as coastal regions.

We also visualised the log probability of dialect terms *hella* (an intensifier mainly used in Northern California) and *yall* (means "you all", used in Southern U.S.) resulting from the Gaussian representation model. As shown in Figure 5, the heatmap matches the expected regions.

## 6 Conclusion

We proposed a continuous representation of location using mixture of Gaussians and applied it to geotagged Twitter data in two different settings: (1) geolocation of social media users, and (2) lexical dialectology. We used `MDN` (Bishop, 1994) in a multilayer neural network as a geolocation model and showed that it outperforms regression models by a large margin. There is also very recent work (Iso et al., 2017) in tweet-level geolocation that shows the effectiveness of MDN.

|  | GEOTEXT | | | | TWITTER-US | | | |
|---|---|---|---|---|---|---|---|---|
|  | regul. | dropout | hidden | $K$ | regul. | dropout | hidden | $K$ |
| Baseline (Regression) | 0 | 0 | (100, 50) | — | $10^{-5}$ | 0 | (100, 50) | — |
| Proposed method (MDN) | 0 | 0.5 | 100 | 100 | $10^{-5}$ | 0 | 300 | 100 |
| Proposed method (MDN-SHARED) | 0 | 0 | 100 | 300 | 0 | 0 | 900 | 900 |

Table 1: Hyper-parameter settings of the geolocation models tuned over development set of each dataset. $K$ is the number of Gaussian components in MDN and MDN-SHARED. Regression has a tanh hidden layer instead of the Gaussian layer. "—" means the parameter is not applicable to the model.

|  | GEOTEXT | | | TWITTER-US | | |
|---|---|---|---|---|---|---|
|  | Acc@161 | Mean | Median | Acc@161 | Mean | Median |
| Baseline (Regression) | 4 | 951 | 733 | 9 | 746 | 557 |
| Proposed method (MDN) | 24 | 983 | 505 | 29 | 696 | 281 |
| Proposed method (MDN-SHARED) | 39 | 865 | 412 | 42 | 655 | 216 |
| CLASSIFICATION METHODS | | | | | | |
| (Rahimi et al., 2015) (LR) | 38 | 880 | 397 | 50 | 686 | 159 |
| (Wing and Baldridge, 2014) (uniform) | — | — | — | 49 | 703 | 170 |
| (Wing and Baldridge, 2014) ($k$-d tree) | — | — | — | 48 | 686 | 191 |
| (Melo and Martins, 2015) | — | — | — | — | 702 | 208 |
| (Cha et al., 2015) | — | 581 | 425 | — | — | — |
| (Liu and Inkpen, 2015) | — | — | — | — | 733 | 377 |

Table 2: Geolocation results over GEOTEXT and TWITTER-US datasets based on Regression, MDN and MDN-SHARED methods. The results are also compared to state-of-the-art classification methods. "—" signifies that no results were reported for the given metric or dataset.

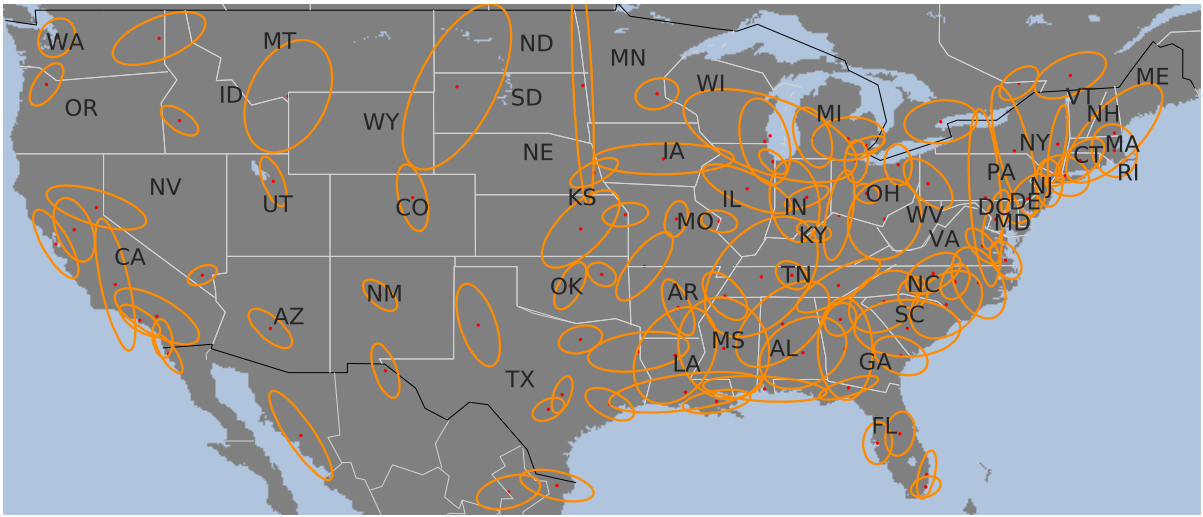

Figure 2: Learned Gaussian representation of input locations over TWITTER-US in the lexical dialectology model. The number of Gaussian components, $K$, is set to 100. The red points are $\mu_k$ and the contours are drawn at $p = 0.01$.

We modified MDN by sharing the parameters of the Gaussian mixtures in MDN-SHARED and improved upon MDN, achieving competetive results with state-of-the-art classification models. We also applied the Gaussian mixture representation to predict dialect words from location, and showed that it

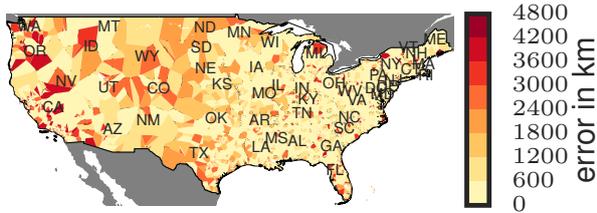

Figure 3: The distribution of geolocation error for `MDN-SHARED` over the development set of TWITTER-US.

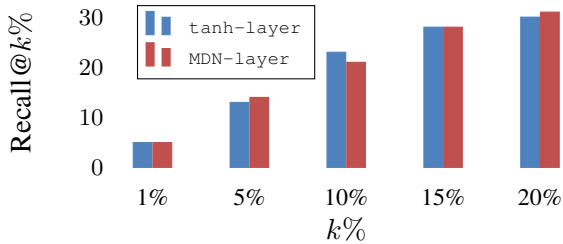

Figure 4: Recall at $k\%$ of 54k vocabulary for retrieving DAREDS dialect words given points within a dialect region.

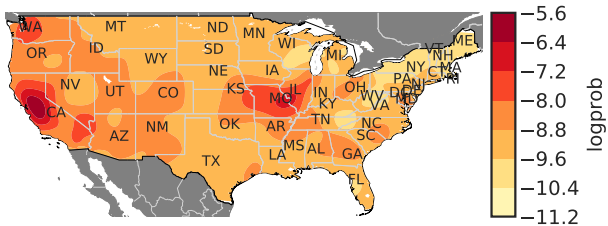

(a) *hella* (an intensifier) mostly used in Northern California, also the name of a company in Illinois.

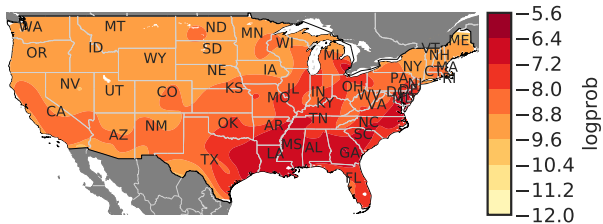

(b) *yall* (means you all) mainly used in Southern U.S.

Figure 5: Log probabilities of terms: (a) *hella* and (b) *yall* in continental U.S.

| New York | Louisiana | Illinois | Pennsylvania |
|---|---|---|---|
| flatbush | kmsl | metra | cdfu |
| lirr | jtfo | osco | jawn |
| reade | wassam | halsted | ard |
| rivington | kmfsl | kedzie | erked |
| mta | ndc | lbvs | cthu |
| nostrand | bookoo | damen | septa |
| stuyvesant | #icantdeal | niu | prussia |
| pathmark | #drove | orland | drawlin |
| bleecker | slangs | cermak | youngbull |
| bowery | daq | uic | prolli |
| macdougal | #gramfam | oms | #ttm |
| broome | gtf | xsport | dickeating |
| driggs | metairie | cta | ctfu |

Table 3: Top terms retrieved from the lexical dialectology model given lat/lon training points within a state ranked by Equation 4.3.

is competitive with simple $\tanh$ activation in terms of both perplexity of the predicted unigram model and also recall at $k$ at retrieving DARE dialect words by location input. Furthermore we showed that the learned Gaussian mixtures have intereting properties such as covering high population density regions (e.g. NYC and LA) with a larger number of small Gaussians, and a smaller number of larger Gaussians in low density areas (e.g. the midwest).

Although we applied the mixture of Gaussians to location data, it can be used in other settings where the input or output are from a continuous multivariate distribution. For example it can be applied to predict financial risk (Wang and Hua, 2014) and sentiment (Joshi et al., 2010) given text. We showed that when a global structure exists (e.g. population centres, in the case of geolocation) it is better to share the global parameters of the mixture model to improve generalisation. In this work, we used the bivariate Gaussian distribution in the `MDN`'s mixture leaving the use of other distributions which might better suit the geolocation task for future research.

## Acknowledgments

We thank the anonymous reviewers for their valuable feedback. This work was supported in part by the Australian Research Council.